\documentclass[sigconf]{acmart}
\AtBeginDocument{%
  }

\setcopyright{acmlicensed}
\copyrightyear{2025}
\acmYear{2025}
\acmConference[MLoG-GenAI@KDD '25]{2025 ACM SIGKDD International Conference on Knowledge Discovery and Data Mining}{Toronto, ON, Canada}
\usepackage{algorithm}
\usepackage{algorithmic}
\usepackage{amsmath, amsfonts}
\usepackage{booktabs, multirow}
\usepackage{colortbl}
\definecolor{customgray}{HTML}{DBDBDB}

\begin{document}

\title{MolX: Enhancing Large Language Models for Molecular Understanding With A Multi-Modal Extension}

% \author{Authors}
% \email{Emails}
% \affiliation{Affiliations\country{}}

\author{Khiem Le\textsuperscript{\rm 1}, Zhichun Guo\textsuperscript{\rm 1}, Kaiwen Dong\textsuperscript{\rm 1}, Xiaobao Huang\textsuperscript{\rm 1}, Bozhao Nan\textsuperscript{\rm 1}, Roshni Iyer\textsuperscript{\rm 2},\\Xiangliang Zhang\textsuperscript{\rm 1}, Olaf Wiest\textsuperscript{\rm 1}, Wei Wang\textsuperscript{\rm 2}, Ting Hua\textsuperscript{\rm 1}, Nitesh V. Chawla\textsuperscript{\rm 1}}
\affiliation{
  \textsuperscript{\rm 1}\institution{University of Notre Dame, IN, USA}\country{}
}
\affiliation{
  \textsuperscript{\rm 2}\institution{University of California, Los Angeles, CA, USA}\country{}
}

\renewcommand{\shortauthors}{Khiem et al.}

\begin{abstract}
Large Language Models (LLMs) with their strong task-handling capabilities have shown remarkable advancements across a spectrum of fields, moving beyond natural language understanding. However, their proficiency within the chemistry domain remains restricted, especially in solving molecule-related tasks. This challenge is attributed to their inherent limitations in comprehending molecules using only common textual representations, i.e. SMILES strings. In this study, we seek to enhance the ability of LLMs to comprehend molecules by equipping them with a multi-modal external module, termed MolX. Instead of directly using SMILES strings to represent a molecule, we utilize specific encoders to extract fine-grained features from both SMILES string and 2D molecular graph representations for feeding into an LLM. A hand-crafted molecular fingerprint is incorporated to leverage its embedded domain knowledge. To establish an alignment between MolX and the LLM’s textual input space, the model in which the LLM is frozen, is pre-trained with a  strategy including a diverse set of tasks. Experimental evaluations show that our proposed method outperforms baselines across downstream molecule-related tasks ranging from molecule-to-text translation to molecular property prediction, with and without fine-tuning the LLM, while only introducing a small number of trainable parameters—0.53\% and 0.82\%, respectively. 
\end{abstract}




\received{June 2025}
\received[revised]{June 2025}
\received[accepted]{June 2025}

\maketitle

\section{Introduction}
Large Language Models (LLMs) have demonstrated impressive performances across a wide array of fields. Extending beyond the boundaries of natural language understanding, LLMs have facilitated various scientific disciplines \cite{taylor2022galactica, telenti2024large}. LLMs have recently been investigated for augmenting research in chemistry as an alternative approach to the traditional supervised learning approach \cite{castro2023large, achiam2023gpt}. 

\begin{figure*}[!t]
    \centering
    \includegraphics[width=1\linewidth]{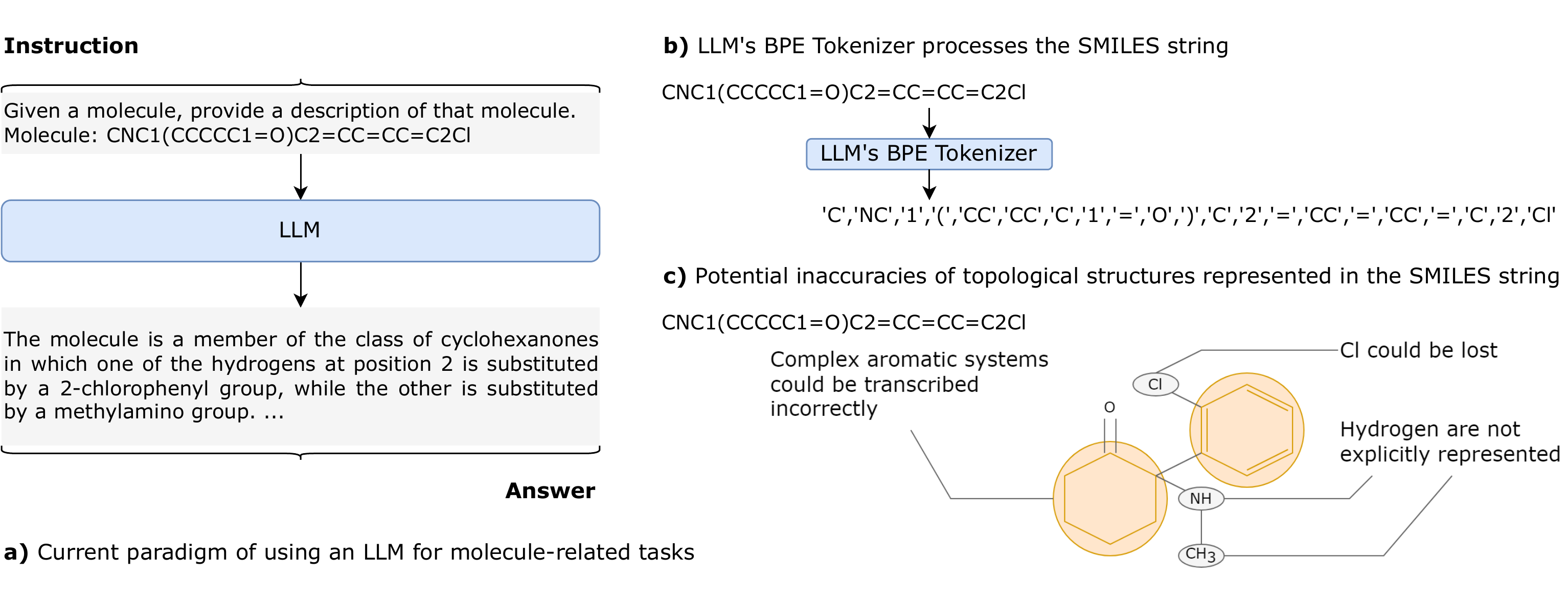}
    \caption{Current paradigm of using an LLM for molecule-related tasks and its issues.}
    \label{Fig1}
\end{figure*}

Despite their strong task-handling capabilities, LLMs still struggle with the chemistry domain, as evidenced by their limited performances on various professional molecule-related tasks \cite{zhao2023scientific, guo2023can}. Llama-2 \cite{touvron2023llama}, performs unsatisfactorily on the molecule-to-text translation tasks such as molecule description generation and IUPAC name generation, providing the correct answer only half as often as supervised learning models. Additionally, such LLM fails to predict molecular properties even using expert-designed prompts. One potential cause of this challenge has been figured out that most existing LLMs represent molecules only by their common textual representations, i.e., SMILES strings \cite{weininger1988smiles}, and process them in a paradigm similar to texts \cite{guo2023can, li2024towards}, as illustrated in Figure \ref{Fig1}a. While convenient, several issues make it challenging for LLMs to comprehend molecules solely by  interpreting SMILES strings. Firstly, LLMs lack an inherent understanding of SMILES strings and blindly treat them as sequences of separate characters relying on their byte-pair encoding tokenizers \cite{sennrich-etal-2016-neural}, which break SMILES strings into smaller pieces in ways that do not represent the chemical principles behind these strings. Without an understanding of these principles, it is difficult for LLMs to capture molecular structure from SMILES strings due to inaccuracies such as incorrect transcription of complex aromatic systems or the absence of hydrogens and other atoms \cite{voinarovska2023yield}, as shown in Figure \ref{Fig1}b and Figure \ref{Fig1}c. 

There have been some early attempts to enhance LLMs for solving molecule-related tasks.  \citet{su2022molecular} employed a GNN-based graph encoder to extract features from the molecule’s 2D molecular graph and directly input such features into the LLM to perform molecule-to-text translation tasks. Developed from that idea, \citet{li2024towards} input features extracted from the 2D or 3D molecular graph into the LLM through an intermediate projector, which is previously aligned with the LLM’s textual input space by a pre-training stage. Although bridging the gap between the 2D or 3D molecular graph and the LLMs, previous approaches are ineffective in using the information contained in a SMILES string, as well as handcrafted molecular descriptors, which have advantages over 2D or 3D molecular graph \cite{david2020molecular, jo2020message}. This might lead to suboptimal performances. Existing methods are only optimized for a limited number of chemistry-related tasks, omitting other crucial tasks such as molecular property prediction. 

\begin{figure*}[!t]
    \centering
    \includegraphics[width=1\linewidth]{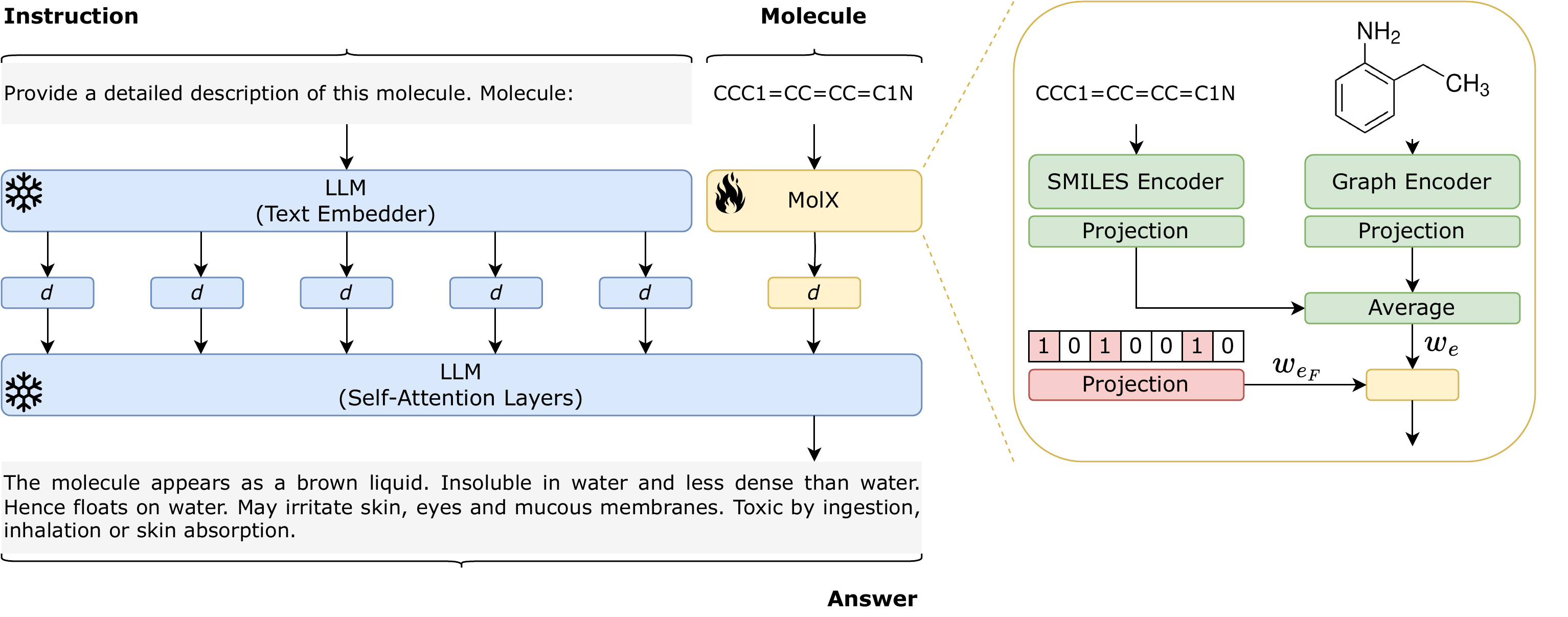}
    \caption{An overview of our proposed method with the main pre-training task.}
    \label{Fig2}
\end{figure*}

In this study, we introduce a novel framework for enhancing LLMs to capture molecules from multiple representations, thus improving their performances on various molecule-related tasks. Our proposed framework consists of two main components which are a multi-modal external module, namely MolX, equipped with the LLMs, and a versatile pre-training strategy for aligning MolX into the LLMs’ textual input space. We first utilize a pre-trained BERT-like \cite{devlin-etal-2019-bert} SMILES encoder to extract features from the SMILES string instead of directly using it to represent the molecule. Because of its initial pre-training, the SMILES encoder works with its tokenizer to capture long-range dependencies encoded in the SMILES string. We simultaneously utilize a pre-trained GNN-based graph encoder to extract features from the molecule’s 2D molecular graph, capturing its topological structures. In addition to features extracted from raw representations, i.e., SMILES string and 2D molecular graph, a handcrafted molecular fingerprint \cite{morgan1965generation} containing domain knowledge is incorporated in a weighted scheme of MolX. Finally, the modelin which the LLM is frozen undergoes  pre-training strategy with a diverse set of tasks, providing the model with information about the molecules. This process provides an alignment between MolX and the LLM’s textual input space. Figure \ref{Fig2} shows an overview of our proposed method. 

Our experimental results demonstrate that the proposed method outperforms baselines by a statistically significant margin on various downstream molecule-related tasks in two different model configurations, with and without fine-tuning the LLM. It is worth noting that MolX can act as a plug-in module to the LLMs for enhancing the performances on molecule-related tasks while fully preserving its general-purpose usage in other domains. 

\newpage
To summarize, our contributions are outlined as follows: 
\begin{itemize}
    \item We introduce a novel framework enhancing LLMs to comprehend molecules, thus improving their performances on various molecule-related tasks. The LLMs are equipped with a multi-modal external module, MolX, to extract features from both SMILES string and 2D molecular graph representations, as well as leverage a handcrafted molecular fingerprint. 
    \item A pre-training strategy including a diverse set of tasks, is applied to establish an alignment between MolX and the LLMs’ textual input space. This process advances the models’ ability of molecular understanding, as well as instruction following. 
    \item Extensive experimental evaluations demonstrate that our proposed method outperforms baselines by a substantial margin on a diverse range of downstream molecule-related tasks in two different model configurations, with and without fine-tuning the LLM. 
\end{itemize}

\section{Related Work}
In this section, we provide a review of the literature related to molecular learning via language modeling and leveraging LLMs for solving molecule-related tasks. 

\subsection{Molecular Learning}
Molecules form the basis of chemistry and molecular learning has been a long-standing problem in cheminformatics \cite{baum2021artificial, xia2023molebert, pei-etal-2023-biot5, pei-etal-2024-biot5}. Traditionally, molecular fingerprints such as Morgan fingerprint \cite{morgan1965generation} or ECFP \cite{rogers2010extended} serve as one of the most important descriptors for molecules, encoding a molecule into a fixed bit string, where each bit indicates the presence of a certain substructure. With the rapid development of language modeling, textual representations such as SMILES strings have become widely used \cite{weininger1988smiles}. Studying molecular property prediction tasks, \citet{wang2019smiles} introduced SMILES-BERT, a BERT-like model \cite{devlin-etal-2019-bert} that is pre-trained with the masked language modeling mechanism (MLM) on a large-scale set of unlabeled molecules. \citet{wang2022chemicalreactionaware} proposed using chemical reactions to assist the pre-training. \citet{ahmad2022chemberta} proposed using auxiliary tasks with more domain relevance for chemistry such as predicting computed properties of molecules, supporting MLM. \citet{irwin2022chemformer} investigated the challenging sequence-to-sequence tasks such as retrosynthesis and introduced Chemformer. \citet{zhong2022root} proposed the root-aligned SMILES (R-SMILES), adopting a tighter representation for those tasks. \citet{edwards-etal-2022-translation} studied molecule-to-text translation tasks and vice versa and proposed MolT5, which is pre-trained with the multi-lingual MLM, considering SMILES strings as a conventional language. \citet{lu2022unified} and \citet{christofidellis2023unifying} presented ChemT5 and Text+ChemT5, unifying all sequence-to-sequence tasks. Several studies \cite{liu2023prediction} demonstrated that fusing the molecule’s 2D molecular graphs with language modeling provides complementary benefits to molecular learning, improving performance on tasks such as molecular property prediction. With rising use across a wide array of fields, including chemistry \cite{castro2023large, achiam2023gpt}, LLMs have emerged as an evolution of the traditional language modeling approach for molecular learning. 

\subsection{LLMs for Molecule-Related Tasks}
Several studies have evaluated LLM applications in chemistry. \citet{castro2023large} explored how well ChatGPT "understands" chemistry by posing five student-level tasks in different subareas of chemistry and noticed moderate performance. \citet{zhao2023scientific} investigated the molecular property prediction task and showed that LLMs relied on memorized information rather than true understanding for making predictions, which limits their applications to new types of molecules required in practical applications. \citet{guo2023can} benchmarked several published LLMs on eight molecule-related tasks. Empirical results reveal that LLMs such as Llama-2 \cite{touvron2023llama} that were widely used at the time typically fail to perform challenging tasks of molecule-to-text translation or predict molecule activity for high-level properties even using expert-designed prompts. A potential reason behind this challenge has been identified that most existing LLMs represent molecules only by their common textual representations, i.e., SMILES strings, which LLMs have a limited understanding of. In response to these findings, \citet{su2022molecular} propose MoMu to enhance LLMs by applying a GNN-based graph encoder to extract features from the molecule’s 2D molecular graph and input such features into the LLM for performing molecule-to-text translation tasks. \citet{li2024towards} proposed 2D and 3D MoLM to leverage an intermediate projector for feeding features extracted from the 2D or 3D molecular graph into the LLM, which is previously aligned with the LLM’s textual input space by a pre-training stage. Despite improvements by bridging the gap between the 2D or 3D molecular graph and the LLMs, the importance of representation  other than SMILES strings such as handcrafted molecular descriptors are underexplored. Existing methods are only optimized for a limited set of molecule-related tasks, how well the enhanced LLMs perform on other tasks such as molecular property prediction is not well understood.

\section{Methodology}
We propose a framework enhancing LLMs to comprehend molecules from multiple representations, consisting of two main components, a multi-modal external module and a novel pre-training strategy. Here, we present the details of these components. 

\subsection{Model Architecture}
The proposed MolX, which is equipped with a base LLM, consists of two key designs: 1) Trainable encoders, focusing on encoding raw representations of a molecule, i.e., SMILES string and 2D molecular graph; 2) A weighted scheme to incorporate a handcrafted molecular fingerprint.

\textbf{Trainable Encoders.} We define a molecule as $m$ and consider $m_S$ and $m_G$ to depict its SMILES string and 2D molecular graph, respectively. While $m_S$ is simply a sequence of ASCII characters, $m_G$ is considered as $m_G = \{\mathcal{V}, \mathcal{E}\}$, where each node in $\mathcal{V}$ indicates an atom and each edge in $\mathcal{E}$ indicates a chemical bond. Also, $\boldsymbol{X} \in \mathbb{R}^{|\mathcal{V}| \times N}$ is the attribute matrix of $m_G$ where $x_n = \boldsymbol{X}[n, :]^T$ is the $N$-dimensional attribute vector of the node $v_n \in \mathcal{V}$.

To encode the SMILES string $m_S$, we adopt a pre-trained BERT-like \cite{devlin-etal-2019-bert} SMILES encoder, ChemBERTa \cite{ahmad2022chemberta}, which is constructed by stacking multiple Transformer layers. ChemBERTa, denoted as $E_S$, is pre-trained on a large-scale set of unlabeled molecules with MLM, enabling it to capture long-range dependencies identified in the SMILES string. An average is taken over outputs of $E_S$ to obtain an embedding vector for $m_S$, which is then projected to the hidden dimension $d$ of the base LLM by a multi-layer perceptron $f_S$: 
\begin{equation}
e_S = f_S(\text{Average}(\{t_i, t_i \in E_S(m_S)\})) \in \mathbb{R}^d. 
\end{equation}

To encode the 2D molecular graph $m_G$, we adopt a pre-trained GNN-based graph encoder, ChemGraphCL \cite{you2020graph}, which is constructed based on an emerging message-passing GNN, GIN \cite{Hou2020Measuring}. ChemGraphCL, denoted as $E_G$, is pre-trained on a large-scale set of unlabeled molecules with a contrastive learning strategy \cite{radford2021learning} and thus able to capture the topological structures of the molecule from its 2D molecular graph. Starting from the initial $x_n$, after multiple layers of message propagation, $E_G$ produces an updated attribute vector $h_n$ for the node $v_n \in \mathcal{V}$. Then, an average is taken over all node-level attribute vectors to obtain an embedding vector for $m_G$, which is projected to the hidden dimension $d$ of the base LLM by a multi-layer perceptron $f_G$: 
\begin{equation}
e_G = f_G(\text{Average}(\{h_n, h_n \in E_G(m_G)\})) \in \mathbb{R}^d. 
\end{equation}

$e_S$ and $e_G$ are then averaged to establish a unified embedding vector $e \in \mathbb{R}^d$.

\textbf{Molecular Fingerprint Incorporation.} Molecular fingerprints are some of the most important descriptors of molecules due to the encoded domain knowledge. While SMILES strings and 2D molecular graphs capture global information about the molecule, molecular fingerprints capture information about the local atomic environments and neighborhoods, explicitly encoding the presence of specific substructures \cite{doi2022screening}. Unfortunately, molecular fingerprints are not often used in deep learning models even though they have been shown to be valuable for specific tasks such as molecular property prediction \cite{xia2024understanding}. We seek to exploit their benefits by incorporating the popular Morgan fingerprint \cite{morgan1965generation} into the unified embedding vector $e$ from trainable encoders described above. RDKit \cite{landrum2013rdkit} is utilized to compute the Morgan fingerprints with a radius of 2 from the molecule $m$, which is then projected to the hidden dimension $d$ of the base LLM by a multi-layer perceptron $f_F$. The incorporation scheme works as follows: 
\begin{equation}
\begin{aligned}
e = w_e \cdot e + w_{e_F} \cdot \text{ }& e_F, \\
\text{where }& e_F = f_F(\texttt{MorganFP}(m)), 
\end{aligned}
\end{equation}
where $w_e$ and $w_{e_F}$ are trainable parameters introduced for providing the model sufficient flexibility to incorporate the Morgan fingerprint into $e$. 

\begin{figure}[!t]
    \centering
    \includegraphics[width=1\linewidth]{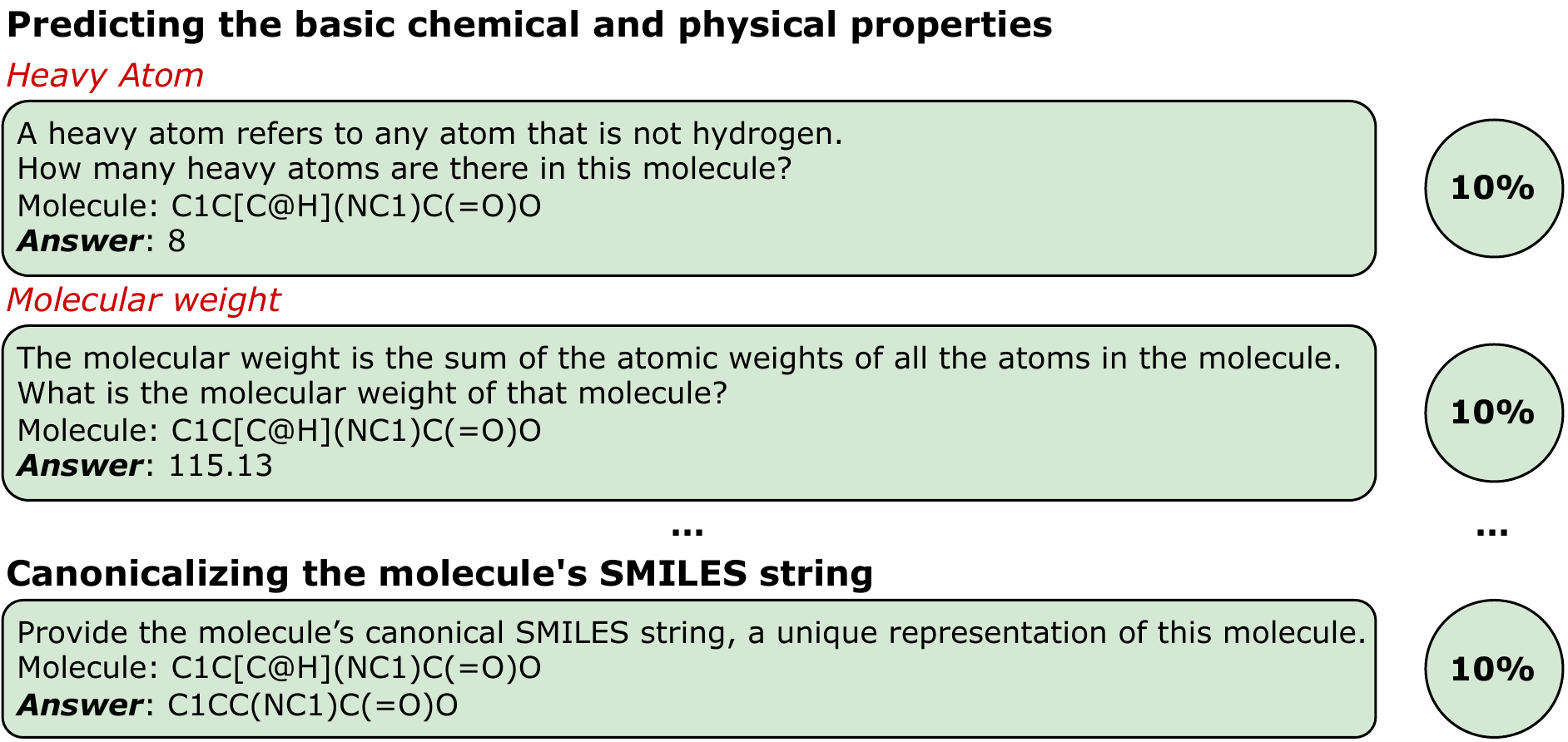}
    \caption{Examples of auxiliary tasks in our instruction-based pre-training strategy.}
    \label{Fig3}
\end{figure}

\begin{table*}[!t]
\centering
\caption{Experimental results for molecule-to-text translation.}
\label{molecule-to-text-translation}
\setlength{\tabcolsep}{2.99pt}
\renewcommand{\arraystretch}{0.90}
\scriptsize

\begin{tabular*}{\linewidth}{@{\extracolsep{\fill}}ll|cccccc|cccccc}
\toprule

\multirow[t]{2}{*}{} &\multirow[t]{2}{*}{Model} &\multicolumn{6}{c}{Description Generation} &\multicolumn{6}{c}{IUPAC Name Generation} \\
& &BLE-2↑ &BLE-4↑ &ROG-1↑ &ROG-2↑ &ROG-L↑ &MET↑ &BLE-2↑ &BLE-4↑ &ROG-1↑ &ROG-2↑ &ROG-L↑ &MET↑ \\
\midrule
\multirow[t]{2}{*}{Infer-only}
&Llama-2-7B                 &03.64 &02.98 &18.28 &04.26 &12.87 &16.21 &05.55 &01.81 &05.40 &00.23 &04.39 &10.30 \\
&Llama-2-7B + \textbf{MolX} &\textbf{08.22} &\textbf{06.40} &\textbf{30.82} &\textbf{21.69} &\textbf{28.94} &\textbf{21.77} &\textbf{10.67} &\textbf{04.76} &\textbf{14.61} &\textbf{01.24} &\textbf{11.47} &\textbf{18.54} \\
\midrule
\multirow[t]{5}{*}{LoRA FT}
&Llama-2-7B                 &27.54 &21.24 &36.50 &21.33 &28.99 &31.69 &51.43 &36.94 &48.54 &20.57 &40.53 &53.38 \\
&Llama-2-7B + MoMu          &27.68 &21.50 &36.76 &21.42 &29.23 &31.86 &51.70 &37.38 &48.89 &20.65 &40.87 &53.66 \\
&Llama-2-7B + MoLM-2D       &27.95 &21.77 &38.66 &22.99 &30.92 &33.69 &52.32 &37.65 &51.77 &21.83 &43.62 &57.10 \\
&Llama-2-7B + MoLM-3D       &29.82 &22.39 &39.12 &23.62 &32.64 &34.34 &55.70 &38.93 &52.03 &22.78 &45.63 &57.84 \\
&Llama-2-7B + \textbf{MolX} &\textbf{31.40} &\textbf{24.25} &\textbf{44.20} &\textbf{28.96} &\textbf{38.76} &\textbf{39.55} &\textbf{56.88} &\textbf{45.01} &\textbf{55.45} &\textbf{30.14} &\textbf{48.19} &\textbf{59.35} \\
%%%%%%%%%%%%%
\rowcolor{customgray}
&LlaSMol-7B                 &26.71 &18.06 &38.75 &22.77 &33.32 &32.63 &49.48 &36.33 &52.38 &28.53 &45.20 &58.48 \\
\rowcolor{customgray}
&ChemDFM-13B                &13.02 &08.30 &20.42 &11.31 &17.93 &18.44 &39.33 &22.83 &37.61 &09.49 &28.68 &45.99 \\
\midrule
\multirow[t]{2}{*}{Full FT}
&MolT5-Large        	    &25.87 &17.28 &34.07 &16.42 &23.41 &28.04 &50.88 &38.69 &45.89 &21.11 &33.03 &44.82 \\
&MolT5-Large + MoMu 	    &26.34 &18.01 &34.75 &16.86 &24.76 &28.73 &51.81 &40.32 &46.81 &21.68 &34.93 &45.92 \\

\bottomrule
\end{tabular*}
\end{table*}

\subsection{Pre-training Strategy}
There is a noticeable misalignment in the latent spaces of MolX and the base LLM where the former encodes molecules while the latter has a textual input space. Therefore a cross-space alignment stage is needed. This is accomplished by feeding the embedding vector from MolX into the LLM as a soft token. We propose to pre-train the MolX-enhanced LLM with a diverse set of tasks including a molecule-to-text translation task, i.e., molecule description generation, accompanied by several auxiliary tasks. It is worth noting that while MolX is trainable, the base LLM is kept frozen during pre-training. This setting maintains the LLM’s inherent generalizability, forcing MolX to produce embedding vectors that are suited in the LLM’s textual input space and can be effectively understood by the LLM to generate accurate answers. This allows the LLM to function normally on general domains by  using MolX as a plug-in module for the handling of molecule-related tasks.

\textbf{Multi-Task Dataset.} To conduct the pre-training, we utilize the pre-train subset of PubChem \cite{li2024towards}, a dataset that contains ~300k molecule-description pairs \footnote{https://pubchem.ncbi.nlm.nih.gov} for the molecule description generation task. By using this task as an objective, MolX is encouraged to produce meaningful embedding vectors, so that the LLM can caption molecules with their substructures and properties accurately, as illustrated in Figure \ref{Fig2}. Although this dataset collected from a reliable source, descriptions in the dataset retain several limitations that might hinder the model’s ability of molecular understanding. The average number of words in the dataset’s descriptions is roughly 20, which is insufficient to describe a molecule. Additionally, some of the dataset’s descriptions are noisy and uninformative \cite{li2024towards}. Therefore, to assist the molecule description task, we design a set of auxiliary tasks including predicting the basic chemical and physical properties of molecules such as the number of heavy atoms or molecular weight. We select a set of 10 low-level properties that are available for easy collection from PubChem and present comprehensive information about the molecules. Further, leveraging the fact that a molecule can be represented by multiple valid SMILES strings \cite{bjerrum2018improving}, we utilize one more special auxiliary task which is canonicalizing the molecule’s SMILES string. This task enhances the model’s understanding of chemical laws behind SMILES strings. To keep the pre-training stage controllable, 10\% of the dataset is used for each auxiliary task. Examples of proposed auxiliary tasks are shown in Figure \ref{Fig3} and details are in Appendix A. 

\textbf{Instruction-based Pre-training.} LLMs tend to exhibit hallucinations in the domain of chemistry \cite{guo2023can}, generating unexpected answers regarding a molecule. Hence, we enrich our pre-training dataset by designing an informative instruction for each task. We then employ instruction-based pre-training \cite{sanh2022multitask, ouyang2022training}, enhancing the model’s ability of instruction following. Formally, we first define $p(.)$ as the textual distribution parameterized by the base LLM. The base LLM is decomposed into two subparts, the text embedder $F_{emb}$ and self-attention layers $F_{att}$, in which the text embedder $F_{emb}$ converts an instruction of a task into a list of $T$ tokens $Z = [z_1, z_2, .., z_T]$. Given a molecule $m$ and its label $y$ for the given task, after the embedding vector $e$ is extracted from MolX, the auto-regressive loss for pre-training is defined as: 
\begin{equation}
\begin{aligned}
\mathcal{L}_{reg} &= -\texttt{log} \hspace{0.2em} p(y|F_{att}(z_1, z_2, .., z_T, e)) \\
&= -\sum_{l=1}^{L} \texttt{log} \hspace{0.2em} p(y_l|F_{att}(z_1, z_2, .., z_T, e), y_1, ..., y_{l-1}), 
\end{aligned}
\end{equation}
where $L$ is the length of the label $y$ for the given task.

\section{Experiments}
In this section, we conduct  experiments on various downstream molecule-related tasks including molecule-to-text translation, molecular property prediction, to demonstrate the effectiveness of our proposed method. Throughout experiments, we utilize Llama-2 \cite{touvron2023llama} with 7B parameters as our base LLM to leverage its text generation capability and internal chemistry knowledge. We consider two different model configurations for the evaluation: I) Inference-only: The model is frozen after pre-training for direct question answering on downstream tasks, evaluating the model’s generalizability without fine-tuning; II) LoRA fine-tuning: The model is fine-tuned on downstream tasks using a parameter-efficient technique, LoRA \cite{hu2022lora}, verifying the model’s adaptability in scenarios where downstream data are available. In addition to direct comparison with previous related works including MoMu \cite{su2022molecular}, as well as 2D and 3D MoLM \cite{li2024towards}, we also compare with competitive supervised learning models in each task. For further reference, we evaluate two recently introduced generalist chemical LLMs derived from Llama-2 \cite{touvron2023llama} that are tailored for molecule-related tasks, i.e., LlaSMol-7B \cite{yu2024llasmol} and ChemDFM-13B \cite{zhao2024chemdfm}. 

The MolX-enhanced LLM is pre-trained with the above tasks in a multi-task learning setting for 5 epochs. AdamW optimizer \cite{loshchilov2018decoupled} is adopted with a weight decay of 0.05 and a learning rate scheduler of a combination of linear warmup with 1000 steps and cosine decay, in which the peak and minimal learning rates are 1e-5 and 5e-6, respectively. The batch size is 12 and the maximal text length is set to be 256. The computation time is 72 hours on 2 A100 GPUs with BFloat16 mixed precision. For experiments on downstream tasks, we consider two different model configurations for the evaluation: I) Inference-only: The model is frozen after pre-training for direct question answering on downstream tasks, evaluating the model’s generalizability without fine-tuning; II) LoRA fine-tuning: The model is fine-tuned on downstream tasks using a parameter-efficient technique, LoRA \cite{hu2022lora}, verifying the model’s adaptability in scenarios where downstream data are available. For LoRA fine-tuning, the model is fine-tuned on train sets of downstream tasks for 50 epochs, using the same settings of optimizer and learning rate scheduler as pre-training. LoRA is applied with the same hyper-parameters as the baselines 2D and 3D MoLM \cite{li2024towards}, factorizing all $*\_proj$ modules of \texttt{LlamaSdpaAttention} and \texttt{LlamaMLP} layers with a rank $r = 8$, $\alpha = 32$, and $dropout = 0.1$. Notably, for all tasks, the loss function employed is the auto-regressive loss as described in Equation (4). We report performances on the test sets selected by the corresponding validation sets. 

\begin{table*}[!t]
\centering
\caption{Experimental results for molecular property prediction.}
\label{molecule-property-prediction}
\setlength{\tabcolsep}{2.99pt}
\renewcommand{\arraystretch}{0.90}
\scriptsize

\begin{tabular*}{\linewidth}{@{\extracolsep{\fill}}ll|ccccccc}
\toprule

\multirow[t]{2}{*}{} &\multirow[t]{2}{*}{Model} &ESOL &FreeSolv &Lipophilicity &HIV &BACE &BBBP &Tox21 \\
& &RMSE↓ &RMSE↓ &RMSE↓ &ACC↑ $|$ F1↑ &ACC↑ $|$ F1↑ &ACC↑ $|$ F1↑ &ACC↑ $|$ F1↑ \\
\midrule
\multirow[t]{2}{*}{Infer-only}
&Llama-2-7B                 &58.719 &357.371 &222.426 &0.135 $|$ 0.129 &0.522 $|$ 0.362 &0.485 $|$ 0.351 &0.090 $|$ 0.084 \\
&Llama-2-7B + \textbf{MolX} &\phantom{5}\textbf{4.929} &\phantom{35}\textbf{9.692} &\phantom{22}\textbf{1.605} &\textbf{0.807 $|$ 0.484} &\textbf{0.530 $|$ 0.524} &\textbf{0.588 $|$ 0.516} &\textbf{0.622 $|$ 0.459} \\
\midrule
\multirow[t]{5}{*}{LoRA FT}
&Llama-2-7B                 &\phantom{5}2.061 &\phantom{35}4.203 &\phantom{22}0.956 &0.960 $|$ 0.610 &0.612 $|$ 0.584 &0.603 $|$ 0.564 &0.740 $|$ 0.578 \\
&Llama-2-7B + MoMu          &\phantom{5}2.112 &\phantom{35}4.214 &\phantom{22}0.998 &0.968 $|$ 0.614 &0.618 $|$ 0.587 &0.612 $|$ 0.574 &0.746 $|$ 0.582 \\
&Llama-2-7B + MoLM-2D       &\phantom{5}1.521 &\phantom{35}3.161 &\phantom{22}0.898 &0.968 $|$ 0.627 &0.631 $|$ 0.599 &0.624 $|$ 0.586 &0.746 $|$ 0.594 \\
&Llama-2-7B + MoLM-3D       &\phantom{5}1.095 &\phantom{35}2.119 &\phantom{22}0.780 &0.968 $|$ 0.640 &0.644 $|$ 0.587 &0.637 $|$ 0.574 &0.746 $|$ 0.606 \\
&Llama-2-7B + \textbf{MolX} &\phantom{5}\textbf{0.967} &\phantom{35}2.371 &\phantom{22}0.808 &\textbf{0.972 $|$ 0.649} &\textbf{0.704 $|$ 0.697} &\textbf{0.666 $|$ 0.650} &\textbf{0.748 $|$ 0.616} \\
%%%%%%%%%%%%%
\rowcolor{customgray}
&LlaSMol-7B                 &\phantom{5}1.871 &\phantom{35}6.047 &\phantom{22}1.361 &0.968 $|$ 0.492 &0.467 $|$ 0.318 &0.529 $|$ 0.346 &0.608 $|$ 0.475 \\
\rowcolor{customgray}
&ChemDFM-13B                &\phantom{5}8.476 &\phantom{35}9.686 &\phantom{22}2.180 &0.952 $|$ 0.534 &0.564 $|$ 0.518 &0.522 $|$ 0.505 &0.677 $|$ 0.529 \\
\midrule
Full FT
&ChemGraphCL  	   	    &\phantom{5}1.231 &\phantom{35}2.951 &\phantom{22}0.822 &0.968 $|$ 0.628 &0.659 $|$ 0.657 &0.638 $|$ 0.629 &0.746 $|$ 0.596 \\
&ChemGraphMVP 		    &\phantom{5}1.091 &\phantom{35}\textbf{2.106} &\phantom{22}\textbf{0.718} &0.971 $|$ 0.630 &0.691 $|$ 0.689 &0.647 $|$ 0.638 &0.747 $|$ 0.597 \\

\bottomrule
\end{tabular*}
\end{table*}

\subsection{Molecule-to-Text Translation}
We first consider the molecule-to-text translation tasks, i.e., molecule description generation and IUPAC name generation. These tasks reflect the general molecular understanding of the model and have crucial applications, enabling humans to gain an overview of a molecule. We conduct experiments on the downstream subset of the PubChem dataset \cite{li2024towards}, which has 15k high-quality molecule-description pairs and is separate from the pre-train one. We opt not to use the CheBI-20 dataset \cite{edwards-etal-2022-translation} because it is also sourced from PubChem and can be viewed as an older version of the used dataset. Following \cite{edwards-etal-2022-translation, li2024towards}, we adopt BLEU-2, BLEU-4, ROUGE-1, ROUGE-2, ROUGE-L, and METEOR as evaluation metrics. 

Table \ref{molecule-to-text-translation} presents experimental results for these tasks across 6 different metrics. Based on the Inference-only results, we observe that the proposed framework significantly enhances the base LLM for direct question answering on both tasks without fine-tuning. In the scenario of LoRA fine-tuning, the MolX-enhanced LLM demonstrates superior performance compared to baselines with the highest scores on all metrics, especially for ROUGE-based and METEOR metrics which might be attributed to the proposed versatile pre-training strategy that provides the model with comprehensive information about the molecules. The approach of fine-tuning the LLM to establish multi-modal models shows better performances than generalist chemical LLMs, i.e., LlaSMol-7B \cite{yu2024llasmol} and ChemDFM-13B \cite{zhao2024chemdfm}, as well as competitive supervised learning models such as MolT5 \cite{edwards-etal-2022-translation} and its MoMu-enhanced one \cite{su2022molecular}. 

\subsection{Molecular Property Prediction}
\begin{figure}[!t]
    \centering
    \includegraphics[width=1\linewidth]{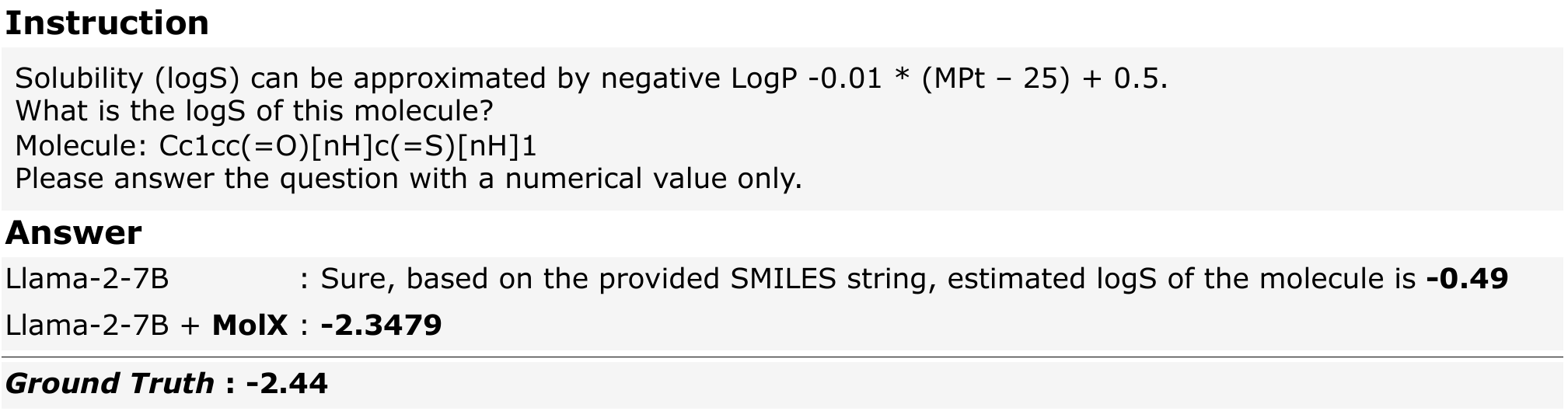}
    \caption{An example of molecular property prediction.}
    \label{Fig4}
\end{figure}

Besides overall understanding, we assess the model’s perception of molecular properties by conducting experiments on the molecular property prediction task. This task involves approximating quantitative attributes such as solubility or determining the activity for assays of a molecule. We employ MoleculeNet dataset \cite{wu2018moleculenet} with 7 different subsets including ESOL, FreeSolv, Lipophilicity, HIV, BACE, BBBP, and Tox21. As evaluation metrics, RMSE is used for regression subsets, and Accuracy and F1 are used for classification, following \cite{yu2024llasmol}. Figure \ref{Fig4} shows an example of this task. 

Experimental results in Table \ref{molecule-property-prediction} show that MolX improves performances of the base LLM in both model configurations. Especially for Inference-only results, MolX remarkably narrows approximation errors. Additionally, MolX enhances the model’s ability of instruction following, generating expected answers without LLMs’s favorite phrases. In addition to LoRA fine-tuned models, we consider ChemGraphCL \cite{you2020graph} which serves as the GNN-based graph encoder in MolX, ensuring an adequate comparison. We observe that the MolX-enhanced LLM achieves the best scores in 6 out of 8 subsets of the MoleculeNet dataset and is the second-best in the other 2. Notably, properties in the MoleculeNet dataset are unseen from the pre-training stage, showing the strong adaptability of our proposed method on unseen downstream tasks. 

\section{Ablation Studies}
Besides the main experiments, we conduct 2 ablation studies on the influence of building components in our proposed framework and verify the agnosticism to the base LLM of MolX where we use another LLM, Mistral \cite{jiang2023mistral}. 

\subsection{Component Impact}

Here we study the influence of building components in our proposed framework. Firstly, we use random initializations for trainable encoders, exploring the possibility of eliminating reliance on robust pre-trained weights. Next, we investigate the contributions of incorporating the Morgan fingerprint, as well as the weighted scheme by removing them from the framework. Moreover, to demonstrate the effectiveness of our versatile pre-training strategy, we discard auxiliary tasks and only use the molecule description generation objective during pre-training. Lastly, by totally skipping the pre-training stage, we aim to understand its alignment impact on the framework. Experiments are conducted on the molecule description generation task on the PubChem dataset \cite{li2024towards} under the LoRA fine-tuning scenario, simultaneously highlighting the proposed framework’s efficiency regarding the number of trainable parameters during pre-training and fine-tuning on downstream tasks. 

Table \ref{influence-of-building-components} shows results for the described ablation study. Firstly, a drop in the performances of MolX without chemical initializations for encoders indicates the role of robust pre-trained weights. Next, while the weighted scheme brings a modest improvement, incorporating the Morgan fingerprint contributed significantly to the performances of MolX. Moreover, without proposed auxiliary tasks, a noticeable decrease in performances can be viewed, especially for ROUGE-based and METEOR metrics, demonstrating their effectiveness in providing the model with comprehensive information about the molecules. Lastly, it is not surprising that the pre-training stage which forms an alignment between MolX and the LLMs’ textual input space, has a large impact. In terms of efficiency, our proposed framework only introduces a small number of trainable parameters, accounting for 0.53\% of the entire parameters during pre-training and 0.82\% with fine-tuning on downstream tasks. 

\begin{table*}[!ht]
\centering
\caption{Added results for molecule description generation.}
\label{influence-of-building-components}
\setlength{\tabcolsep}{2.99pt}
\renewcommand{\arraystretch}{0.90}
\scriptsize

\begin{tabular*}{\linewidth}{@{\extracolsep{\fill}}ll|cccccccc}
\toprule
\multirow[t]{2}{*}{} &\multirow[t]{2}{*}{Model} &\multicolumn{2}{c}{\# Trainable Params} &\multicolumn{6}{c}{Description Generation} \\
& &Pre-training &Downstream &BLE-2↑ &BLE-4↑ &ROG-1↑ &ROG-2↑ &ROG-L↑ &MET↑ \\
\midrule
\multirow[t]{6}{*}{LoRA FT}
& Llama-2-7B + MolX w/o ChemInit          & 36.1M (0.53\%) & 56.6M (0.82\%) &30.21 &22.67 &43.64 &28.80 &38.47 &38.43 \\
& Llama-2-7B + MolX w/o \texttt{MorganFP} & 23.5M (0.35\%) & 44.0M (0.64\%) &29.33 &22.01 &42.37 &27.96 &37.35 &37.31 \\
& Llama-2-7B + MolX w/o WeightedInc       & 36.1M (0.53\%) & 56.6M (0.82\%) &31.13 &24.01 &44.16 &28.50 &38.56 &39.34 \\
& Llama-2-7B + MolX w/o Auxiliaries       & 36.1M (0.53\%) & 56.6M (0.82\%) &30.71 &23.06 &40.29 &24.33 &33.62 &35.37 \\
& Llama-2-7B + MolX w/o Pre-training      & 00.0M (0.00\%) & 56.6M (0.82\%) &28.79 &22.36 &38.23 &22.28 &30.40 &33.13 \\
& Llama-2-7B + \textbf{MolX}              & 36.1M (0.53\%) & 56.6M (0.82\%) &\textbf{31.40} &\textbf{24.25} &\textbf{44.20} &\textbf{28.96} &\textbf{38.76} &\textbf{39.55} \\

\bottomrule
\end{tabular*}
\end{table*}

\subsection{LLM Agnosticism}
Since we propose MolX as a plug-in module to the LLMs, here we verify its agnosticism to the base LLM by conducting experiments with another LLM, Mistral \cite{jiang2023mistral} with 7B parameters. Experiments are conducted on the molecule description generation task on the PubChem dataset \cite{li2024towards} under the LoRA fine-tuning scenario with the same experimental settings for a fair comparison. 

\begin{figure}[!ht]
    \centering
    \includegraphics[width=1\linewidth]{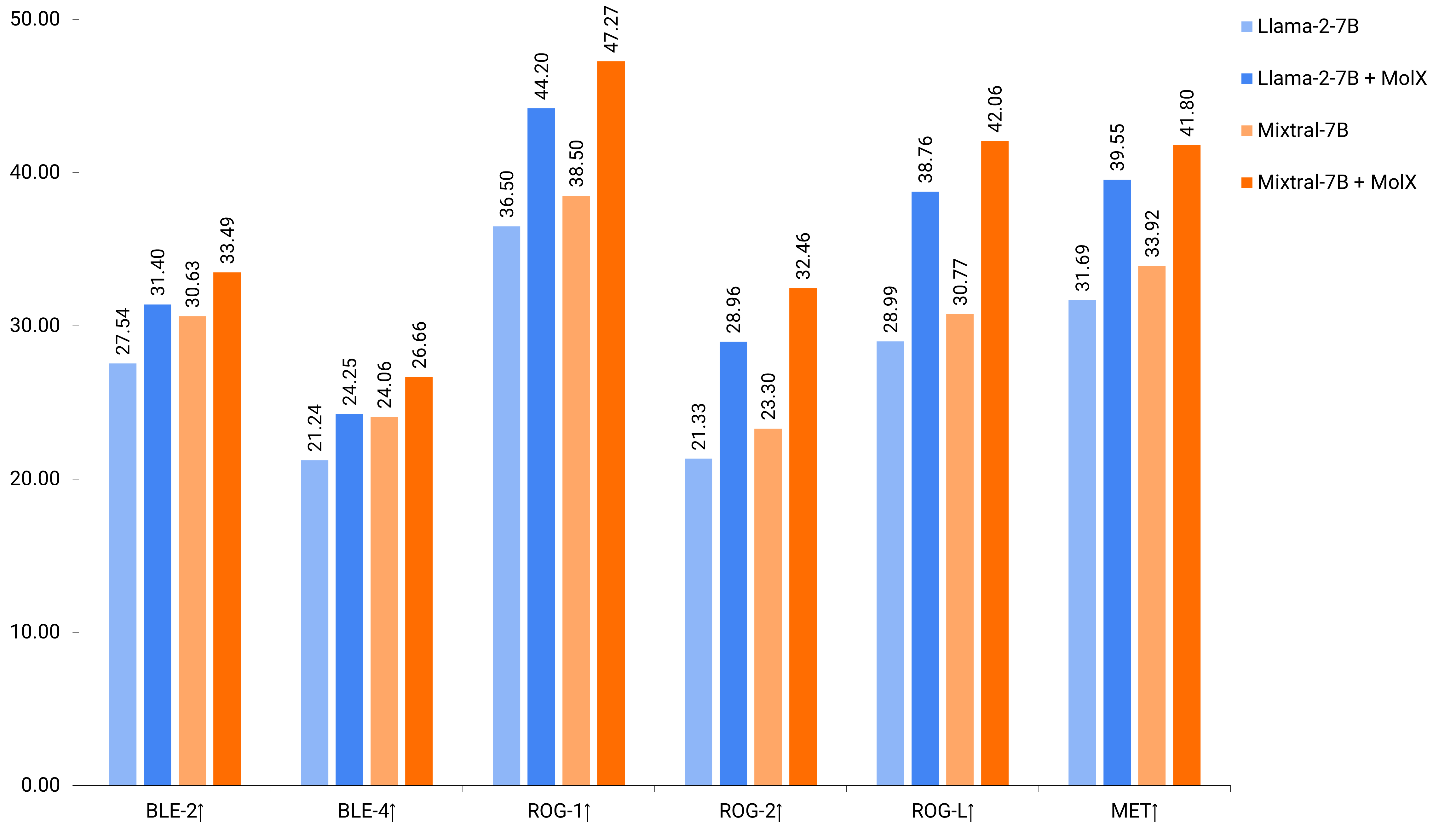}
    \caption{Added results for molecule description generation.}
    \label{FigA1}
\end{figure}

Figure \ref{FigA1} shows results for the described ablation study. First, we can observe that MolX consistently improves performances of Mistral \cite{jiang2023mistral} across 6 different metrics, as proved with Llama-2 \cite{touvron2023llama} earlier. Additionally, as Mistral \cite{jiang2023mistral} outperforms Llama-2 \cite{touvron2023llama}, the MolX-enhanced Mistral \cite{jiang2023mistral} outperforms the MolX-enhanced Llama-2 \cite{touvron2023llama} across 6 different metrics. This verifies the compatibility of MolX with different base LLMs and also proves that it could be applied to stronger LLMs released in the future. 

\section{Discussion}
First, Table \ref{efficiency} demonstrates that our proposed method is designed to be more efficient than most baselines while giving superior performances.

Here we discuss the limitations of our work and future directions. Firstly, we are aligning MolX into the LLM via a soft token, which is simple but effective. Although we are aware of advanced cross-space alignment techniques such as Q-Former \cite{li2023blip}, we opt not to employ them since they require a large number of high-quality molecule-description pairs and an extra pre-training stage, leading to high computational costs. A better alignment technique tailored for molecule-related tasks needs to be explored. Moreover, throughout experiments, we show the limitations of current generalist chemical LLMs, therefore, a novel generalist chemical LLM enhanced with MolX should be developed. LLMs also have been demonstrated to have intriguing abilities like In-context Learning \cite{NEURIPS2020_1457c0d6} or Chain-of-Thought \cite{wei2022chain}. Leveraging these abilities for molecule-related tasks is a potential direction. 

\begin{table}[!t]
\centering
\caption{Numbers of trainable parameters in experiments.}
\label{efficiency}
\setlength{\tabcolsep}{2.99pt}
\renewcommand{\arraystretch}{0.90}
\scriptsize

\begin{tabular*}{\linewidth}{@{\extracolsep{\fill}}ll|rr}
\toprule

\multirow[t]{2}{*}{} &\multirow[t]{2}{*}{Model} &\multicolumn{2}{c}{\# Trainable Params} \\
& &Pre-training↓ &Downstream↓ \\
\midrule
\multirow[t]{5}{*}{LoRA FT}
&Llama-2-7B                 &   0.0M (0.00\%) &  20.5M (0.30\%) \\
&Llama-2-7B + MoMu          &   2.0M (0.00\%) &  22.5M (0.30\%) \\
&Llama-2-7B + MoLM-2D       & 120.0M (1.74\%) & 120.0M (1.74\%) \\
&Llama-2-7B + MoLM-3D       & 120.0M (1.74\%) & 120.0M (1.74\%) \\
&Llama-2-7B + \textbf{MolX} &  36.1M (0.53\%) &  56.6M (0.82\%) \\
%%%%%%%%%%%%%
\rowcolor{customgray}
&LlaSMol-7B                 &   0.0M (0.00\%) & 113.2M (1.64\%) \\
\rowcolor{customgray}
&ChemDFM-13B                &    13B (100.\%) &   13B (100.\%) \\
\midrule
Full FT
&MolT5-Large                &   0.0M (0.00\%) & 780.1M (100.\%) \\
&MolT5-Large + MoMu         &   2.0M (0.00\%) & 782.1M (100.\%) \\

\bottomrule
\end{tabular*}
\end{table}

\section{Conclusion}
In this paper, we propose a novel framework enhancing LLMs to comprehend molecules, thus, improving their performances on molecule-related tasks. The LLMs are equipped with a multi-modal external module, MolX, which is aligned with their textual input space using a versatile pre-training strategy. Experimental evaluations show that our proposed method consistently outperforms baselines across downstream molecule-related tasks ranging from molecule-to-text translation to molecular property prediction, with and without fine-tuning the LLM, while only introducing a small number of trainable parameters—0.53\% and 0.82\%, respectively. 

\section{Acknowledgement}
This work was supported by the National Science Foundation (CHE–2202693) through the NSF Center for Computer Assisted Synthesis (C-CAS).

% \clearpage
\bibliographystyle{ACM-Reference-Format}
\bibliography{sample-base}

% \clearpage
\appendix

\begin{figure*}[!ht]
    \centering
    \includegraphics[width=1\linewidth]{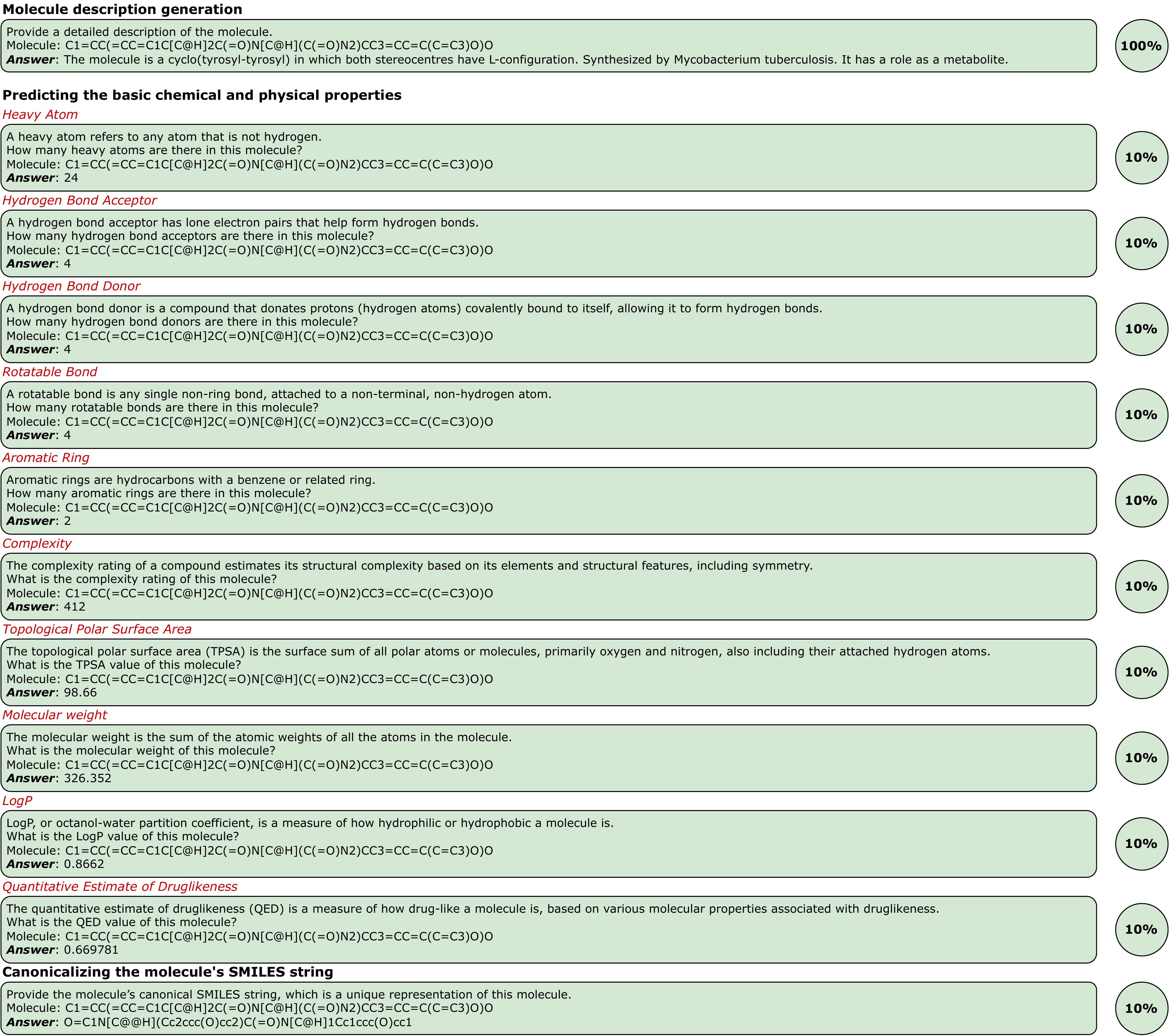}
    \caption{Examples of all pre-training tasks in our instruction-based pre-training strategy.}
    \label{FigA3}
\end{figure*}

\section{Methodology}

Here we elaborate the pre-training strategy by describing all proposed pre-training tasks. The molecule description generation task serves as the main task, accompanied by a couple of auxiliary tasks. We select a set of 10 low-level properties that present comprehensive information about the molecules. We use one more special auxiliary task which is canonicalizing the molecule’s SMILES string. Examples of these tasks and their instructions are illustrated in Figure \ref{FigA3}.

\end{document}